\documentclass{ceurart}

\usepackage{graphicx}
\usepackage{booktabs}
\usepackage{multirow}
\usepackage{url}
\definecolor{finalformatblue}{RGB}{220,230,255}

\copyrightclause{Copyright for this paper by its authors.
Use permitted under Creative Commons License Attribution 4.0
International (CC BY 4.0).}

\conference{Workshop on Quality of Knowledge Graphs at ESWC 2026, May 11, 2026, Dubrovnik, Croatia}

\usepackage[utf8]{inputenc}

\DeclareUnicodeCharacter{2212}{\ensuremath{-}}
\DeclareUnicodeCharacter{2248}{\ensuremath{\approx}}
\DeclareUnicodeCharacter{2013}{-}
\DeclareUnicodeCharacter{2014}{-}
\DeclareUnicodeCharacter{00A0}{ }

\begin{document}

\title{An Explainable Header-Centric Framework for Large-Scale Semantic Table Interpretation and Data Quality Assessment}

\address{Curtin University, Kent Street, Bentley, Perth, Western Australia, 6102, Australia}

\author[1]{Marcelo Valentim Silva}[
  email=marcelo.valentim@bcb.gov.br,
  orcid=0000-0003-4592-2968
]
\cormark[1]
\cortext[1]{Corresponding author.}

\author[2]{Hannes Herrmann}[
  email=hannes.herrmann@curtin.edu.au,
  orcid=0000-0002-3160-4988
]

\author[3]{Valerie Maxville}[
  email=v.maxville@curtin.edu.au,
  orcid=0000-0001-8842-7364
]

\begin{abstract}
Knowledge Graph (KG) quality is influenced not only by downstream graph validation, but also by the quality of the tabular metadata used prior to integration. In metadata-only Semantic Table Interpretation (STI), where cell values are unavailable, noisy, or unsuitable for use, column headers become a critical source of semantic evidence. This is directly relevant to metadata quality, provenance-aware traceability, and quality-aware preparation for knowledge-graph integration.

We present an explainable, header-centric framework for metadata-only Column Type Annotation (CTA) and Data Quality Assessment (DQA). The framework maps headers to a compact layer of 39 interpretable \textit{FinalFormat} types using curated lexical resources, while preserving token-level traceability through \textit{SourceKeywords}. Each assigned type activates predefined validation rules grounded in a structured taxonomy of Data Quality Issues (DQI), producing cell-level DQI instances such as missing data, duplicates, domain violations, wrong data type, and temporal mismatch. These detections are aggregated into \textit{HeadersIQ}, a lightweight and intentionally unweighted data source-level data quality metric relevant to the SemTab 2024 IsGold? track and related KG quality assessment work, such as KG2Tables. Illustrative case studies from real data sources show that identifier semantics simultaneously trigger completeness and uniqueness constraints, whereas bounded numerical formats expose domain violations.

The framework was evaluated across heterogeneous benchmarks, including UCI, Prague, Kaggle, VizNet/Sato, SOTAB, T2Dv2, and the SemTab 2024 Metadata-to-KG track, comprising around 120,000 header columns. This large-scale evaluation demonstrates broad practical coverage across noisy real-world metadata, while a parallel KG-mapping pathway supports alignment to DBpedia and Schema.org. On the SemTab 2024 Metadata-to-KG track, the official GT-strict evaluation was modest. Still, a blinded diagnostic audit indicates that many mismatches reflect benchmark granularity, aliasing, and ontology-selection effects rather than wholly implausible header-centric predictions. We report this audit as diagnostic evidence on the structure of disagreements, not as revised benchmark performance. Overall, the paper presents a reusable and auditable workflow for metadata-driven semantic annotation, data source-level quality monitoring, and KG-oriented benchmark diagnosis. This positions the framework as a practical contribution to quality-aware tabular preparation and the broader goal of sustaining trustworthy knowledge graph ecosystems. 
\end{abstract}

\maketitle

\section{Introduction}

In recent years, the rapid expansion of open data repositories and domain-specific tabular data sources has made Table Understanding (TU) a central challenge for semantic integration and interoperability with knowledge graphs ~\cite{ref36}. Most real-world tables remain poorly annotated or lack explicit semantic types, making semantic integration difficult and prone to errors.  Within TU, Semantic Table
Interpretation (STI) addresses this need by linking tabular data to ontologies and knowledge graphs such as DBpedia \cite{ref4} and Schema.org \cite{Guha2015SchemaorgEO}. In this setting, the Column Type Annotation (CTA) subtask assigns semantically meaningful types to table columns, supporting semantic alignment and enabling richer metadata for downstream data quality assessment and knowledge-graph integration. 

\begin{figure}[htbp]
  \centering
  \includegraphics[width=0.8\linewidth]{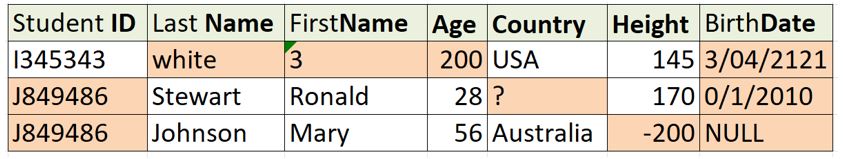}
  \caption{\centering Example of Bad Data/Dirty Dataset.}
  \label{fig:fig1}
\end{figure}

Figure~\ref{fig:fig1} illustrates a realistic "dirty" dataset with duplicates, missing values, and obvious typographical problems. In our framework, such problems are detected and explained through the semantic types inferred from the terms that appear in bold in the column headers.

\subsection{Historical context and renewed interest}

Influential research such as \textbf{Sherlock (2019) \cite{ref13}} discouraged header-centric features, arguing that matching-based approaches (i) recognise only a limited range of types and (ii) are not robust to malformed or context-dependent data, particularly when relying on dictionary or regular-expression matching. This influenced community practice, where unrecognised headers often fall back to primitive types.  

Recent findings challenge this view. 
A survey by \textbf{Liu et al. (2023) \cite{ref23}} reports that competitive CTA systems increasingly exploit header signals: \textbf{\textit{“The table header often directly explains the contents of the column”}} and that using header information can help predict column types or properties more efficiently.

Since 2019, the Semantic Web Challenges on Tabular Data to Knowledge Graph Matching (SemTab Challenges) \cite{ref16,ref17,ref8,ref11,ref7,ref36} have reinforced this shift. In particular, the SemTab 2024 edition \cite{ref11} introduced a Metadata-to-KG track in which cell values could not be used, reflecting practical concerns such as privacy, sensitive data, and large-scale processing. SemTab 2024 also included the \textbf{IsGold? track \cite{ref1}}, which further emphasized automated \textbf{data quality assessment}. \textbf{KG2Tables (Abdelmageed et al., 2025) \cite{ref2}} also highlighted data quality as an open gap (diversity, annotation accuracy/clarity, structural coherence). These developments further reinforce the relevance of scalable and interpretable metadata-based methods.

In particular, AdaTyper (2023) \cite{ref15}, from the Sherlock/Sato group, prioritises header analysis before cell-based fallback, further supporting the value of header semantics.

Taken together, these developments motivate an approach that mitigates the limitations noted by Sherlock \cite{ref13}. Dynamic dictionary expansion increases coverage beyond the limited type range typically associated with matching-based systems. Normalisation and abbreviation mappings improve robustness to noisy, malformed, and lexically variable metadata that would otherwise defeat dictionary-based matching. In addition, the \textit{\textbf{FinalFormat}} layer extends semantic coverage beyond simple primitive categories, although context-dependent ambiguity remains a challenging case.

\subsection{Approach and Contributions}

This work extends a previous header-centric semantic typing and data quality framework \cite{ref37} into a broader framework for metadata-only Semantic Table Interpretation (STI) and Data Quality Assessment (DQA). Its central premise is that column headers can provide the primary semantic evidence needed for interpretable column typing and downstream quality assessment, especially when cell values are noisy, unavailable, sensitive, or unsuitable for use.

The framework follows a header-centric workflow in which headers are normalised and expanded through curated lexical resources. After preprocessing, the pipeline branches into two parallel pathways. In the first, semantic type detection maps headers to interpretable semantic types in the \textit{\textbf{FinalFormat}} layer. These assignments activate targeted data quality checks grounded in a structured taxonomy of \textbf{Data Quality Issues}. In the second, a separate KG-mapping pathway aligns preprocessed headers to DBpedia and Schema.org for KG-oriented interpretation and benchmark comparison. 

Traceability is preserved through \textit{\textbf{SourceKeywords}}, and detected violations can be aggregated through \textit{\textbf{HeadersIQ}} for data source-level quality monitoring. The framework is then evaluated across heterogeneous benchmarks to assess practical coverage and robustness under noisy metadata conditions. Here, \textit{\textbf{FinalFormat}} denotes the paper’s compact semantic type layer, \textit{\textbf{SourceKeywords}} are the matched header tokens that justify each assignment, and \textit{\textbf{HeadersIQ}} is the resulting data source-level summary metric derived from activated DQA checks.

\vspace{1em}

\textbf{This paper makes four main contributions:}

\textbf{1.	Framework.} We propose an explainable header-centric framework for metadata-only semantic typing that links inferred semantic types to rule-grounded data quality assessment, while a parallel KG-mapping pathway supports KG-oriented interpretation and benchmark comparison.

\textbf{2.	Semantic layer and traceability.} We define a compact \textit{\textbf{FinalFormat}} abstraction layer, supported by curated lexical resources and token-level traceability mechanisms to enable interpretable and auditable semantic assignments.

\textbf{3.	Large-scale evaluation.} We provide a broad empirical study across heterogeneous benchmarks to examine the scalability, coverage, and robustness of header-centric annotation under noisy real-world metadata conditions.

\textbf{4.	Quality monitoring and benchmark diagnosis.} We introduce \textit{\textbf{HeadersIQ}} for data source-level quality monitoring and show how header-centric outputs can support diagnostic analysis of benchmark and ground-truth limitations in KG-oriented evaluation.

Taken together, these contributions position the framework as a scalable, explainable, metadata-only approach to semantic type annotation, provenance-aware traceability, and KG-oriented analysis.

In addition, ongoing work includes a \textbf{Statistical Audit} designed to provide correctness-oriented evidence for \textit{\textbf{FinalFormat}} assignments beyond the coverage-focused indicators reported in this paper. A replication package, including appendices and implementation details, is publicly available at \cite{ref38}. Subsequent references to supplementary materials in the paper refer to this package.

\section{Related Work}

\subsection{Column Type Annotation (CTA) Paradigms}

In this paper, the relevant setting is metadata-only CTA, where semantic inference relies on headers and related metadata rather than cell values.

Following Liu et al. \cite{ref23}, CTA systems can be grouped into four paradigms: 

(i) heuristic dictionary-based systems that match header tokens against controlled vocabularies or KG labels, such as Wang et al. \cite{ref44}, C2 \cite{ref19}, MAGIC \cite{ref40}, and Alobaid et al. \cite{ref3}; 

(ii) heuristic and iterative methods that propagate constraints or distribute semantic cues across table elements, including TableMiner+ \cite{ref47}, CSV2KG \cite{ref41}, T2K \cite{ref34} and the MTab family \cite{ref29,ref30,ref31}; 

(iii) feature-based machine learning approaches that use engineered features over values and/or headers, for example, Limaye et al. \cite{ref22}, and Mulwad et al. \cite{ref27,ref28}; and 

(iv) deep learning and embedding-based methods, including Sherlock \cite{ref13}, Sato \cite{ref45}, TURL \cite{ref10}, Doduo \cite{ref42}, RECA \cite{ref39}, and DAGOBAH \cite{ref6}, which exploit representations learned from large table corpora. Only a small subset of systems (12 in Liu’s \textit{T}0 category) deal with headers, and even fewer operate without access to cell values, leaving an underexplored region in the CTA landscape and motivating our header-centric, metadata-driven design. A broader comparative analysis, including additional recent models, is provided in \textbf{Appendix A \cite{ref38}}.

\newpage

\subsection{Attribute-Level Data Quality Dimensions}
\label{sec:attribute_dq_dimensions}

Building on a previous framework \cite{ref37}, which established the link between semantic type assignments and executable validation rules, the present work extends this integration with a formally grounded DQI (Data Quality Issues) taxonomy and large-scale empirical evaluation.

Attributes (columns) constitute the primary semantic units of tabular data. Classical data quality literature identifies several core dimensions associated with attribute-level assessment \cite{BatiniScannapieco2016,DasuJohnson2003}. These include:

\begin{itemize}
\item \textbf{Accuracy}: the closeness between a recorded value and the corresponding real-world entity or state.
\item \textbf{Completeness}: the degree to which expected values are present (absence of missing tuples or values).
\item \textbf{Consistency}: the absence of violations of semantic or structural rules (e.g., type constraints, integrity constraints).
\item \textbf{Uniqueness}: the absence of unintended duplicates that represent the same real-world entity.
\item \textbf{Timeliness}: the validity of temporal information with respect to expected time bounds or recency.
\end{itemize}

Many data quality problems can be interpreted as violations of one or more of these dimensions. However, without semantic interpretation of attribute labels, automated profiling tools often fail to determine which constraints should apply. In this work, semantic type detection provides the missing link between metadata interpretation and dimension-aware validation.

\subsection{Data Quality Issues (DQIs) and Taxonomic Grounding}
\label{sec:dqi_taxonomy}

To operationalise dimension-aware validation, we adopt the structured taxonomy of 22 Data Quality Issues (DQIs) compiled by Visengeriyeva and Abedjan (2020) \cite{VisengeriyevaAbedjan2020}. This taxonomy maps concrete error patterns to specific Data Quality Dimensions.

Examples of DQIs particularly relevant to attribute-level validation include

\begin{itemize}
\item \textbf{DQI \#1 – Missing Data} (Accuracy, Completeness): null values, blank entries, disguised placeholders.
\item \textbf{DQI \#5 – Extraneous Data} (Consistency, Uniqueness): additional unexpected content within attribute values.
\item \textbf{DQI \#9 – Duplicates} (Uniqueness): repeated tuples or values representing the same entity.
\item \textbf{DQI \#13 – Temporal Mismatch} (Accuracy, Timeliness): invalid or structurally incorrect temporal values.
\item \textbf{DQI \#15 – Domain Violation} (Accuracy): values violating semantic constraints of the attribute.
\item \textbf{DQI \#17 – Wrong Data Type} (Consistency): values violating declared or inferred data type.
\item \textbf{DQI \#19 – Uniqueness Violation} (Uniqueness): explicit violation of identifier-level uniqueness constraints.
\end{itemize}

Although certain DQIs (e.g., referential integrity violations across relations) require cross-table analysis and fall outside the scope of this study, a substantial subset can be deterministically activated through attribute-level semantic interpretation.

\newpage

\section{Methodology}

\begin{figure}[htbp]
  \centering
  \includegraphics[width=0.7\linewidth]{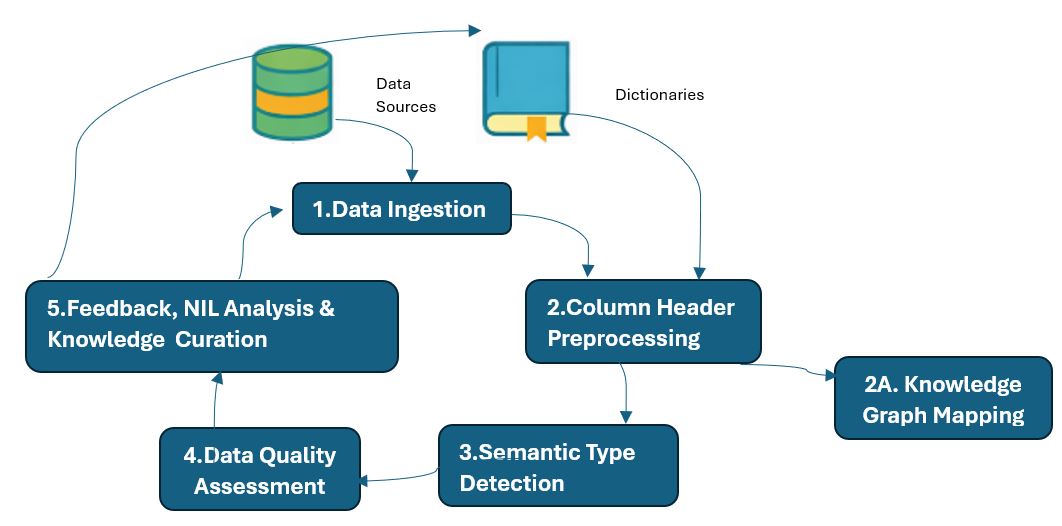}
  \caption{\centering Pipeline for Semantic Type Detection, Knowledge Graph Mapping, and Data Quality Assessment.}
  \label{fig:fig2}
\end{figure}

Figure 2 summarises the end-to-end workflow. After preprocessing, the pipeline branches into two parallel pathways: semantic type detection, which supports downstream data quality assessment, and knowledge graph mapping, which supports KG-oriented interpretation and benchmark comparison. NIL analysis feeds back into dictionary curation.

\subsection{Data Ingestion}
The ingestion stage serves as the entry point for tabular data from various sources, ranging from relational databases to datasets distributed in file-based formats such as CSV files, text files, and spreadsheets, particularly those obtained from large repositories such as UCI \cite{ref24} or Kaggle \cite{ref18}. Because these sources vary widely in structure, ingestion handles multiple file formats and extracts initial metadata such as column headers, optional descriptions, and, when available, indicators of primary or foreign keys. Basic validation filters incomplete or malformed tables, and character encodings are normalised to ensure consistent downstream processing. The resulting metadata forms the basis for dictionary expansion and header preprocessing. 

A key component of the framework is a pair of curated dictionaries: 

(i) a \textbf{\textit{formats dictionary}} (formats\_dictionary.txt \cite{ref38}) that maps normalised header tokens (synonyms, variants, and domain-specific terms) to 39 \textit{\textbf{FinalFormat}} types (see Table 2, Section 3.3), and 

(ii) an \textbf{\textit{abbreviations dictionary}} (abbreviations\_dictionary.txt  \cite{ref38}) that expands shorthand forms into terms registered in the \textbf{\textit{formats dictionary}}. 

\vspace{1em}
\textbf{\textit{Formats Dictionary} Creation and Growth}. 

The \textbf{\textit{formats dictionary}} \cite{ref38} was initially derived from header tokens in \textbf{earlier work \cite{ref37}} and later expanded with DBHeaders \cite{ref9}, derived from Web Data Commons \cite{ref21} over more than 90 million tables in the Web Table Corpus \cite{ref5}. Through iterative NIL-driven review over almost 119,000 analysed columns, it grew to roughly 2,800 entries and now supports 39 \textbf{\textit{FinalFormat}} types. Table 1 illustrates representative mappings. These mappings ensure that semantically related or ambiguous headers are consistently normalised to interpretable \textit{\textbf{FinalFormat}} types, such as \textit{name, city, binary, categorical}, and \textit{postalcode}. The coverage results reported in Sections 4.2 and 4.4 reflect the dictionary snapshot available when each benchmark was processed. 

\begin{table}[htbp]
\centering
\caption{Example mappings from terms to \textit{\textbf{FinalFormats}} (in \textbf{\textit{formats dictionary}} file \cite{ref38})}
\label{tab:finalformats_mappings}
\begin{tabular}{ll ll}
\toprule
\textbf{Term} & $\rightarrow$ \textbf{\textit{FinalFormat} type} & \textbf{Term} & $\rightarrow$ \textbf{\textit{FinalFormat} type} \\
\midrule
id                & $\rightarrow$ \textit{IDcolumn}       & author            & $\rightarrow$ \textit{name}      \\
day and time      & $\rightarrow$ \textit{datetime}       & website           & $\rightarrow$ \textit{URLformat} \\
boolean           & $\rightarrow$ \textit{binary}         & day of week       & $\rightarrow$ \textit{weekday}   \\
qualitative       & $\rightarrow$ \textit{categorical}    & address           & $\rightarrow$ \textit{street}    \\
height            & $\rightarrow$ \textit{numerical>=0} & postal code   & $\rightarrow$ \textit{postalcode} \\
birthplace        & $\rightarrow$ \textit{city}           & internet protocol & $\rightarrow$ \textit{IPformat}  \\
territory         & $\rightarrow$ \textit{state}          & cellphone         & $\rightarrow$ \textit{phone}     \\
\bottomrule
\end{tabular}
\end{table}

\newpage

\textbf{\textit{Abbreviation Dictionary} Creation and Expansion.} 

\begin{figure}[htbp]
  \centering
  \includegraphics[width=0.8\linewidth]{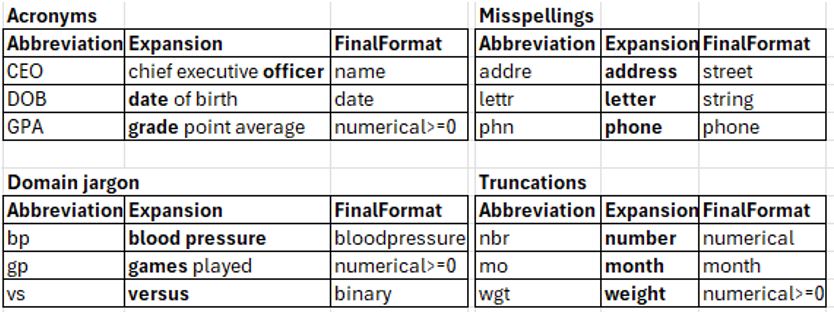}
  \caption{\centering Abbreviation categories and their expansions.}
  \label{fig:fig3}
\end{figure}

To complement the \textbf{\textit{formats dictionary}}, we maintain a dedicated \textbf{\textit{abbreviations dictionary}} \cite{ref37}, \cite{ref38} that contains over 1,000 mappings from shorthand or compressed forms to their expanded variants that map to terms present in the \textbf{\textit{formats dictionary.}} This expansion is crucial for decoding terse headers (e.g., \textit{“DOB” → \textbf{date} of birth → date} or \textit{“pct” → percentage}).
Figure~\ref{fig:fig3} illustrates representative abbreviation categories, covering acronyms, misspellings (e.g., \textit{“addre” → address, “phn” → phone,} drawn partly from NameGuess \cite{ref46}), domain jargon and truncations. In this figure, the bold terms in the Expansion column correspond to the \textit{\textbf{SourceKeywords}} that trigger the \textit{\textbf{FinalFormat}} assignment. In every case, at least one term in the expanded form links to a \textit{\textbf{FinalFormat}} type; for example, \textit{CEO → chief executive \textbf{officer} → name.} Here the \textit{\textbf{SourceKeyword}} is \textit{\textbf{officer}}, which is linked to \textit{\textbf{FinalFormat}} \textit{name} in the \textbf{\textit{formats dictionary}}.

Like the main \textbf{\textit{formats dictionary}}, the \textbf{\textit{abbreviation dictionary}} also grows iteratively through NIL analysis and expert review, supporting continual adaptation as new data sources and types of noise are encountered.

\textbf{Alignment with the SemTab 2025 Challenge Priorities.} This combined strategy of dictionary growth and abbreviation expansion aligns with several key priorities highlighted in SemTab 2025 \cite{ref7}, particularly robust alias/acronym resolution, noise resilience, NIL detection for unmappable headers, and ambiguity handling in tabular metadata. 

\subsection{Column Header Preprocessing}

Preprocessing converts raw column headers into actionable semantic annotations using the \textit{\textbf{FinalFormat}} type system. The main stages are the following.

\textbf{1.	Metadata Extraction and Normalisation.}  Extract header terms, normalise case and punctuation, and split compound or camelCase tokens to standardise heterogeneous naming conventions.
\newpage

\textbf{2.	Abbreviation and Variant Handling.} Expand over 1,000 abbreviations and spelling variants, enabling robust treatment of noisy, truncated, or misspelled headers commonly found in real-world data.

\textbf{3. \textit{SourceKeywords} Extraction.} Extract and record the dictionary-matched trigger token(s) from the cleaned header and optional description (after abbreviation expansion) that led to the \textit{\textbf{FinalFormat}} assignment, providing token-level traceability. 

For example, in \textit{has context, “has”} is recorded as the \textbf{\textit{SourceKeyword}} that triggers the \textbf{\textit{FinalFormat} binary}; similarly, \textit{“bp”} expands to \textit{“blood pressure”}, which is recorded as the \textbf{\textit{SourceKeyword}} triggering the assignment to \textbf{\textit{FinalFormat}} \textit{bloodpressure}.

\subsection{Semantic Type Detection}

The \textit{\textbf{FinalFormat}} system provides the semantic backbone of the framework, defining 39 interpretable types, listed in Table 2, that sit between coarse atomic data types (e.g., string, integer, date) and fine-grained ontology classes in DBpedia or Schema.org. This layer is expressive enough to support meaningful data-quality checks while remaining simple and explainable. 

\vspace{1em}
\textbf{Definition and Scope. }

\textit{\textbf{FinalFormat}} types cover categories such as:

•	\textbf{Numerical}: generic numeric, non-negative, and bounded numerical types.

•	\textbf{Geographical}: \textit{city, country, postalcode, state, street}.

•	\textbf{Temporal}: \textit{date, datetime, day, hour, month, time, week, weekday, year}.

•	\textbf{Categorical}: Keywords such as “class”, “status”, “type”, etc.

•	\textbf{Name}: Keywords such as “director”, “actor”, “publisher”, or “firstName”.

•	\textbf{Identifiers}: \textit{IDcolumn} (e.g.“ISBN”, “SSN”), \textit{URLformat, IPformat, phone.}

•	\textbf{Binary}: Boolean fields or fields identified through cues (“has”).

•	\textbf{Textual}: descriptive, free text, assigned to \textit{\textbf{FinalFormat}} string.

\vspace{-0.4em}
{
\small
\centering
\setlength{\tabcolsep}{6pt}
\renewcommand{\arraystretch}{1.0}
\captionof{table}{The 39 \textit{\textbf{FinalFormats}}. Blue shading indicates bounded numerical \textit{\textbf{FinalFormats}}.}
\label{tab:finalformats}
\begin{tabular}{lll}
\hline
\textbf{FinalFormat} & \textbf{FinalFormat} & \textbf{FinalFormat} \\
\hline
\cellcolor{finalformatblue}acidity        & \cellcolor{finalformatblue}heartrate     & \cellcolor{finalformatblue}percentage \\
\cellcolor{finalformatblue}age            & \cellcolor{finalformatblue}hour          & \cellcolor{finalformatblue}pH \\
\cellcolor{finalformatblue}alkalinity     & IDcolumn                                 & phone \\
\cellcolor{finalformatblue}angle          & IPformat                                 & postalcode \\
binary                                    & \cellcolor{finalformatblue}latitude      & \cellcolor{finalformatblue}saltness \\
\cellcolor{finalformatblue}bloodpressure  & \cellcolor{finalformatblue}longitude     & state \\
categorical                               & modelname                                & street \\
city                                      & \cellcolor{finalformatblue}money         & string \\
country                                   & month                                    & time \\
date                                      & name                                     & URLformat \\
datetime                                  & \cellcolor{finalformatblue}normalized    & week \\
day                                       & numerical                                & weekday \\
E-mailformat                              & numerical>=0                         & year \\
\hline
\end{tabular}
\par
}
\vspace{1em}
\textbf{Matching and annotation.}
Semantic type assignment is performed through deterministic rule-based matching.
Exact, substring and fuzzy matching are applied to assign \textit{\textbf{FinalFormat}} types using the curated \textbf{\textit{Formats}} and \textbf{\textit{Abbreviations dictionaries.}} \textit{\textbf{SourceKeywords}} record the supporting header or description tokens for explanation and auditing.

\newpage

\subsection{Knowledge Graph Mapping}
Each column was linked to the most semantically aligned property in DBpedia \cite{ref4} and Schema.org \cite{Guha2015SchemaorgEO}. This mapping stage operates as a parallel pathway after preprocessing, placing header-centric metadata into a knowledge-graph vocabulary space for KG-oriented interpretation and benchmark comparison.

Figure 4 illustrates real mappings from the KG-mapping pathway, showing header normalisation, abbreviation expansion, dictionary matching, \textit{\textbf{FinalFormat}} assignment, \textit{\textbf{SourceKeyword}} extraction, and ontology alignment with DBpedia and Schema.org. The examples highlight how preprocessing improves semantic alignment, as abbreviated or compressed headers such as Wt, Yr, ISO, OS, NOC, and obp are expanded before mapping. Several cases produce accurate ontology matches, such as weight, year, Organization, operatingSystem, and percentage, while others reveal limitations in ontology coverage or granularity, including NIL results in Schema.org and the more specific copyrightYear returned for Yr. 

These cases show that KG-based annotation can benefit from abbreviation expansion, but semantic accuracy still depends on the available vocabulary and modelling choices of the target ontology. Importantly, such KG-mapping limitations do not affect \textit{\textbf{FinalFormat}} assignment or downstream DQA activation. After preprocessing, semantic type detection and KG mapping proceed as parallel pathways: \textit{\textbf{FinalFormat}} assignment supports DQA, whereas KG mapping is used for ontology-oriented interpretation and benchmark comparison.

\begin{figure}[htbp]
  \centering
  \includegraphics[width=1.0\linewidth]{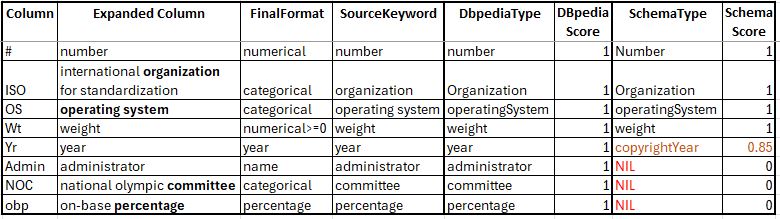}
  \caption{\centering Examples of \textit{\textbf{FinalFormat}} assignments and KG mappings.}
  \label{fig:fig4}
\end{figure}

Overall, \textit{\textbf{FinalFormat}} delivers semantically richer classifications than atomic data types, while remaining more compact and operational than ontology mappings, thereby enhancing interpretability and data-quality profiling.

\subsection{Data Quality Assessment}
\label{sec:semantic_rule_activation}

As introduced in earlier work \cite{ref37}, each inferred \textit{\textbf{FinalFormat}} activates one or more predefined sets of data quality validation functions. These checks include, for example, uniqueness enforcement for identifier formats, bounded-range validation for numerical formats, and structural validation for dates and categorical fields. 

In the present study, rule activation is grounded in the DQI taxonomy (Section~\ref{sec:dqi_taxonomy}). Each validation function yields cell-level DQI instances, which are aggregated into \textit{\textbf{HeadersIQ}} (Subsection~\ref{sec:HeadersIQ}).

Representative examples include:

\begin{itemize}
\item \texttt{check\_numerical\_ge\_zero}: enforces non-negative constraint (DQI \#1, \#15, \#17).
\item \texttt{check\_numerical\_between}: validates bounded numeric ranges (DQI \#1, \#15, \#17).
\item \texttt{check\_id\_attributes}: validates identifier properties (DQI \#1, \#9, \#15, \#17, \#19).
\item \texttt{check\_date}: validates temporal structure and bounds (DQI \#1, \#6, \#13, \#15).
\item \texttt{check\_postal\_code}: validates structural constraints (DQI \#1, \#15, \#17).
\item \texttt{check\_email\_format}: enforces email syntax validity (DQI \#1, \#15).
\end{itemize}

For bounded numerical \textit{\textbf{FinalFormats}}, \texttt{check\_numerical\_between} applies a predefined interval associated with the inferred semantic type. For example, a column assigned to \textit{\textbf{FinalFormat}} \textit{age} is validated against the adopted interval 0--130, while \textit{pH} is validated against 0--14. The validator, therefore, does not require additional per-dataset input at runtime; it uses the interval linked to the assigned \textit{\textbf{FinalFormat}}. These bounds are maintained as configurable domain parameters linked to \textit{\textbf{FinalFormat}} types rather than hard-coded properties of individual datasets, so they can be revised independently as domain requirements or governance policies evolve.

This mapping ensures that rule activation is not heuristic, but is grounded in a predefined taxonomy of violations aligned with classical data quality theory. Each detected violation is recorded as a cell-level DQI instance.

\subsubsection{\textit{\textbf{HeadersIQ}}: A Metric for Data/Information Quality Monitoring}
\label{sec:HeadersIQ}

\textit{\textbf{HeadersIQ}} was developed as a lightweight, unweighted summary metric for monitoring data source-level quality. It reflects the proportion of evaluated cells with no detected violations under the activated data quality rules. The name \textit{\textbf{HeadersIQ}} reflects both its foundation in column headers and its focus on Information Quality (IQ).
\paragraph{Formal definition.}
Let $N_c$ be the number of columns, $N_r$ the number of rows, and $\mathcal{X}$ the set of evaluated cells. Since all cells are evaluated, the total number of evaluated cells is
\[
T = |\mathcal{X}| = N_c \times N_r.
\]

For each $x \in \mathcal{X}$, define
\[
v(x) =
\begin{cases}
1, & \text{if at least one activated data quality rule detects a violation in cell } x,\\
0, & \text{otherwise.}
\end{cases}
\]

Then
\[
V = \sum_{x \in \mathcal{X}} v(x)
\]

denotes the number of distinct cells with one or more detected violations. 

The \textit{\textbf{HeadersIQ}} score is defined as
\begin{equation}
\mathrm{\textit{\textbf{HeadersIQ}} } = \left(1 - \frac{V}{T}\right) \times 100
\end{equation}

It produces a normalised 0-100 score representing the proportion of evaluated cells for which no violations were detected under the activated rules.

\subsection{Feedback, NIL Analysis \& Knowledge Curation}

NIL-labelled headers are reviewed through a controlled, evidence-based human-in-the-loop feedback cycle that evolves with each benchmark iteration.

\textbf{NIL Token improvement process. }
All columns with \textit{\textbf{FinalFormat}} labelled as NIL follow this protocol:

• Aggregate and rank all unmatched tokens by frequency.

• Filter out meaningless or unstable entries.

• Manually review high-frequency candidates for domain relevance.

• Integrate only validated and interpretable terms into the \textbf{\textit{formats}} and \textbf{\textit{abbreviations dictionaries.}}
\newpage
\section{Results and Coverage Analysis}

This section presents a comprehensive evaluation of the header-centric semantic type detection framework across a diverse suite of tabular data benchmarks, including both classical STI datasets and large, real-world sources. We report coverage for \textit{\textbf{FinalFormat}} assignments and Knowledge Graph mappings (DBpedia and Schema.org), and complement quantitative results with qualitative interpretability evidence via \textit{\textbf{SourceKeywords}}.

Sherlock \cite{ref13} highlighted two main limitations of matching-based, header-dependent systems: they tend to recognise only a limited range of types, and they lack robustness when confronted with noisy or context-dependent attributes. The evaluation presented in this chapter directly addresses both points. Broad lexical coverage is demonstrated through the \textbf{\textit{dictionaries (formats}} and \textbf{\textit{abbreviations})} and the 39 \textit{\textbf{FinalFormat}} types. Robustness is demonstrated through consistently high assignment rates across heterogeneous and noisy repositories, such as Kaggle and VizNet, where normalisation, abbreviation expansion, \textit{\textbf{SourceKeywords}}, and NIL-driven feedback enhance resilience under conditions where matching-based systems have previously been shown to fail.

\subsection{Data Sources and Benchmarking Scope}

Scalability, robustness, and practical coverage were evaluated across seven heterogeneous benchmarks (Table 3). The evaluation spans structured repositories (\textbf{UCI \cite{ref24}, Prague Relational Learning Repository \cite{ref26})}, large collections of web tables (\textbf{Sato} \cite{ref45}, \textbf{VizNet} \cite{ref12}), user-generated real-world tables (\textbf{Kaggle \cite{ref18})} and KG-grounded benchmarks (\textbf{SOTAB} \cite{ref20}, \textbf{T2Dv2} \cite{ref35}).

\begin{table}[htbp]
\centering
\caption{Structural Statistics of Benchmarks.     T2Dv2 Domains (*) correspond to DBpedia classes.}
\label{tab:benchmarks_stats}
\begin{tabular}{lrrrr}
\toprule
\textbf{Benchmark} & \textbf{Domains} & \textbf{Data Sources} & \textbf{Columns} & \textbf{Columns per Data Source} \\
\midrule
UCI    & 6   & 50    & 922     & 18.4 \\
Prague & 9   & 50    & 520     & 10.4 \\
Sato   & N/A & 2,254 & 4,587   & 2.0  \\
VizNet & N/A & 11,199& 74,915  & 6.7  \\
Kaggle & 17  & 3,364 & 49,727  & 14.8 \\
SOTAB  & N/A & N/A   & 737     & N/A  \\
T2Dv2  & 39(*) & 237   & 1,172   & 4.9  \\
\bottomrule
\end{tabular}
\end{table}

\textbf{Universe definitions.}

\textbf{Universe A} - All seven benchmarks. 

\textbf{Universe B} - Union of unique header labels extracted from Kaggle and VizNet, comprising 118,639 unique headers (after deduplicating identical header strings across both sources). Used to study distribution, long-tail behaviour and bounded-numerical coverage at scale.

\subsection{Semantic Type Detection: Coverage and Diversity}

We measure three aspects of semantic coverage across all benchmarks:

•	\textit{\textbf{FinalFormat}} Types: number detected in each data source group.

•	\textit{\textbf{SourceKeywords}}: unique, normalized header tokens extracted across all.

•	DBpedia/Schema.org Types: number of distinct ontology classes/properties successfully linked from column headers to the respective ontologies after normalisation and abbreviation expansion.

Table~4 summarises coverage. VizNet and Kaggle yield higher \textit{\textbf{SourceKeywords}} counts due to scale; notably, Kaggle attains full \textit{\textbf{FinalFormat}} type coverage (39/39 types).

\begin{table}[htbp]
\centering
\caption{Coverage of \textit{\textbf{FinalFormat}}, \textit{\textbf{SourceKeywords}}, and KG Types}
\label{tab:coverage_finalformat_sourcekeywords_kgtypes}
\begin{tabular}{lrrrr}
\toprule
\textbf{Benchmark} & \textbf{\textit{FinalFormat} Types} & \textit{\textbf{SourceKeywords}} & \textbf{DBpedia Types} & \textbf{Schema.org Types} \\
\midrule
UCI    & 33 & 248   & 162  & 161  \\
Prague & 23 & 139   & 136  & 129  \\
Sato   & \textbf{15} & \textbf{76}    & \textbf{76 }  & 63   \\
VizNet & 35 & \textbf{1,792} & 1,045& 1,010\\
Kaggle & \textbf{39} & \textbf{1,920} & 1,046& 1,138\\
SOTAB  & 23 & 349   & 246  & 684  \\
T2Dv2  & 21 & 234   & 213  & 172  \\
\bottomrule
\end{tabular}
\end{table}

A key result is that \textbf{only 15 core \textit{\textbf{FinalFormats}}} deterministically cover the entire \textbf{76-label Sherlock/Sato space} in many-to-one mappings. Two extra labels from Sherlock do not appear in Sato (\textit{File size and Team name}). 

The consolidation preserves distinctions important for data quality (e.g., \textit{IDcolumn, date, postalcode}), and redundant semantic variants (e.g., creator-type labels) merge naturally into compact categories, such as \textit{name} or \textit{categorical}.

Within \textbf{Universe B (Kaggle + VizNet)}, these 15 formats alone cover \textbf{$\sim$80\%} of the more than 118,000 headers. The distribution is heavily concentrated in a few operationally critical types (\textit{categorical, numerical>=0, IDcolumn, string, numerical}). The complete list of these 15 types and their mappings to Sherlock/Sato is shown in \textbf{Appendix B \cite{ref38}}.

Beyond the 15 core types, \textbf{Universe B} exhibits an additional 24 \textit{\textbf{FinalFormats}}, primarily bounded numerical values and domain-specific formats such as \textit{percentage}, chemical (\textit{ph, acidity}), health (\textit{heartrate}), geography (\textit{latitude/longitude}), among others. These collectively represent $\sim$6\% of headers, \textbf{totalling over 7,000 in this universe, which remain substantial in absolute terms,} and enable finer-grained Data Quality Issues (DQI) checks not supported in the Sherlock/Sato taxonomy. The complete list is also reported in \textbf{Appendix B \cite{ref38}}.

Across \textbf{Universe B, NIL is approximately 13 percent (thus Valid is $\sim$87 percent)}. Here, NIL denotes headers that did not match any \textit{\textbf{FinalFormat}} after normalisation and abbreviation expansion. 

\paragraph{\textbf{Scalability.}} The modest increase of 80\% from $\sim$2,100 dictionary entries (UCI phase \cite{ref37}) to $\sim$3,800 entries supported a growth of $\sim$103 times in practical coverage, increasing valid assignments from $\sim$1,000 to $\sim$103,000 headers. Details are provided in Table 5.

\begin{table}[htbp]
\centering
\caption{Scalability evidence}
\label{tab:scalability_evidence}
\begin{tabular}{l l r l}
\toprule
\textbf{Phase} & \textbf{Dictionary size} & \textbf{Headers evaluated} & \textbf{Valid headers assigned} \\
\midrule
UCI baseline~\cite{ref37} & 1,800 + 300 $\sim$ 2,100 & $\sim$1,000 & $\sim$1,000 (near perfect) \\
Kaggle+VizNet     & 2,800 + 1,000 $\sim$ 3,800 & 118,639    & $\sim$103,000 ($\sim$87\%) \\
\bottomrule
\end{tabular}
\end{table}

\textbf{Bounded numerical formats. Appendix C \cite{ref38}} lists all adopted bounds used in this study. These intervals support finer-grained DQI checks and help disambiguate numeric families within the \textit{\textbf{FinalFormat}} taxonomy.  

\subsection{NIL Token Improvement Process}

A qualitative review of the top 100 NIL cases in \textbf{Universe B} shows that most unresolved headers are extremely short codes or symbols (e.g., \textit{no, so, ga, po}) that carry little or no semantic information and therefore fall outside any meaningful \textit{\textbf{FinalFormat}} mapping.

A smaller subset of cases (e.g., \textit{streak, medal, designation, msrp}) exhibits domain-specific or plausible abbreviations. These are added only when their meaning is stable, and the mapping is interpretable.

These observations confirm that the remaining NIL proportion mainly reflects low-information tokens or highly specialised terminology rather than systematic gaps. The complete ranked list of 100 NIL tokens is included in \textbf{Appendix D \cite{ref38}}.

\subsection{Validity, NIL Rates, and Robustness}

Table 6 reports NIL counts and Valid percentages for (i) \textit{\textbf{FinalFormat}} assignments and (ii) KG mappings to DBpedia and Schema.org across \textbf{Universe A} (all seven benchmarks).

\textbf{Definition of \textbf{NIL}} and \textbf{Valid\%: NIL} indicates the absence of a match after all normalisation and expansion steps. For all systems, \textbf{Valid\%} is computed as ((\textbf{Total Columns − NIL}) / \textbf{Total Columns}) × 100. A \textbf{Valid\%} column in \textit{\textbf{FinalFormat}} corresponds to a successful mapping to one of the 39 defined semantic types after all dictionary expansions.
For DBpedia and Schema.org, \textbf{Valid\%} measures the fraction of columns whose normalised header (after the same cleaning and abbreviation expansion steps) could be aligned to at least one ontology property/class by the KG mapping component described earlier. This KG alignment is derived directly from header to ontology label similarity and is therefore independent of the \textit{\textbf{FinalFormat}} assignment.

\begin{table}[htbp]
\centering
\caption{NIL Counts and Valid Percentages for \textit{\textbf{FinalFormat}}, DBpedia and Schema.org}
\label{tab:nil_valid_rates}
\begin{tabular}{l r rr rr rr}
\toprule
\multirow{2}{*}{\textbf{Benchmark}} & \multirow{2}{*}{\textbf{Total Columns}} &
\multicolumn{2}{c}{\textit{\textbf{FinalFormat}}} &
\multicolumn{2}{c}{\textbf{DBpedia}} &
\multicolumn{2}{c}{\textbf{Schema.org}} \\
\cmidrule(lr){3-4} \cmidrule(lr){5-6} \cmidrule(lr){7-8}
 &  & \textbf{NIL} & \textbf{Valid\%} & \textbf{NIL} & \textbf{Valid\%} & \textbf{NIL} & \textbf{Valid\%} \\
\midrule
UCI    & 922    & 12    & 98.7 & 433   & 53.0 & 413   & 55.2 \\
Prague & 520    & 0     & 100.0& 94    & 81.9 & 93    & 82.1 \\
Sato   & 4{,}587& 0     & 100.0& 0     & 100.0& 423   & 90.8 \\
VizNet & 74{,}915& 9{,}435& 87.4 & 21{,}244& 71.6 & 23{,}795& 68.2 \\
Kaggle & 49{,}727& 7{,}274& 85.4 & 16{,}546& 66.7 & 17{,}467& 64.9 \\
SOTAB  & 737    & 5     & 99.3 & 163   & 77.9 & 7     & 99.1 \\
T2Dv2  & 1{,}172& 267   & 77.2 & 305   & 74.0 & 469   & 60.0 \\
\bottomrule
\end{tabular}
\end{table}

\textbf{Key Observations.}

•	\textit{\textbf{FinalFormat}} consistently achieves the highest Valid percentages across the benchmarks, reflecting the broader lexical coverage provided by the curated dictionaries and the traceability supported by \textit{\textbf{SourceKeywords}}.

•	Sato obtains 100\% Valid for \textit{\textbf{FinalFormat}}, indicating full coverage of this benchmark and compatibility with the Sherlock/Sato label space. 

•	Sato also reaches 100\% Valid for DBpedia, since its annotated semantic labels correspond directly to DBpedia classes, demonstrating strong compatibility between this benchmark and the ontology vocabulary.

•	By contrast, KG mapping coverage is notably lower for UCI and Kaggle, which is consistent with their more heterogeneous headers, greater use of domain-specific vocabulary, and cases in which suitable ontology properties or classes are unavailable or similarity-based alignment yields NIL.

\subsection{Handling Ambiguous Headers}

Beyond straightforward headers, the system also resolves complex, abbreviated, and domain-specific cases through the combined effect of normalisation, abbreviation expansion and \textit{\textbf{SourceKeywords}} extraction. These mechanisms enable the framework to interpret signals such as \textit{“is”, “has” (binary} cues), \textit{“range” (categorical} context), \textit{“birth place”} (geographical semantics) or medical abbreviations such as \textit{“trestbps” (bloodpressure)}. This avoids misclassification in cases where multiple semantic cues coexist or where the lexical form is highly compressed.

These challenging cases demonstrate the system’s ability to surpass surface-level matching and produce interpretable, rule-aligned \textit{\textbf{FinalFormats}}, even in noisy or unconventional metadata. A detailed description of the rule-based preprocessing, normalisation, abbreviation expansion and KG annotation steps, together with an extended set of worked examples showing how \textit{\textbf{SourceKeywords}} and \textit{\textbf{FinalFormats}} are assigned to complex headers, is provided in \textbf{Appendix E (Table G) \cite{ref38}}.

\subsection{SemTab 2024 Metadata-to-KG Evaluation}

\textbf{Ground-truth note}. As detailed in \textbf{Appendix F \cite{ref38}}, the \textbf{SemTab-2024 Metadata-to-KG ground truth (GT)} exhibits typical large-benchmark issues, including duplicates, label-granularity mismatches and occasional errors. We therefore reported results in two layers: 
\begin{enumerate}
  \item The \textbf{official GT-strict scores}, which enable head-to-head comparison with other systems, and
  \item Blinded \textbf{internal diagnostics for our own outputs}, which characterise disagreement patterns under header-centric metadata-only evidence.
\end{enumerate}

\textbf{Official (GT-strict) results.} Using the organisers’ ground truth and scripts, the system correctly annotated 63/141 columns (Hit@1 = 0.45, 95\% CI ≈ [0.36, 0.53]) and 67/141 (Hit@5 = 0.48, 95\% CI ≈ [0.40, 0.56]). Baselines for Adwan \cite{ref43} and for CVA/ MetaLinker \cite{ref25} were taken from the SemTab-2024 official report \cite{ref11}. These head-to-head figures appear in \textbf{Panel A of Table 7}.

\begin{table}[htbp]
\centering
\caption{Comparative analysis on the SemTab 2024 Metadata-to-KG track}
\label{tab:semtab2024_comparative_panels}

\fbox{%
\begin{minipage}{0.97\linewidth}
\centering

\textbf{Panel A - Official (GT-strict)}\\[0.6em]

\begin{tabular}{|l|c|c|}
\hline
\textbf{System} & \textbf{Hit@1} & \textbf{Hit@5} \\
\hline
Ours (GT-strict) & 0.45 [0.36, 0.53] & 0.48 [0.40, 0.56] \\
\hline
Adwan (Vandemoortele et al., 2024) \cite{ref43} & 0.75 & 0.92 \\
\hline
CVA / MetaLinker (Martorana et al., 2024) \cite{ref25} & 0.55 & 0.70 \\
\hline
\end{tabular}

\vspace{1.2em}

\textbf{Panel B - Diagnostic (ours only)}\\[0.6em]

\begin{tabular}{|l|c|c|}
\hline
\textbf{Category} & \textbf{Support@1} & \textbf{Support@5} \\
\hline
\textbf{B1:} Strict + C.1 (GT-Confirmed) & 0.45 [0.36, 0.53] & 0.48 [0.40, 0.56] \\
\hline
\textbf{B2:} B1 + C.2 (GT-Refinement) & 0.75 [0.68, 0.82] & 0.78 [0.70, 0.84] \\
\hline
\textbf{B3:} B2 + C.3 (Alternative) & 0.79 [0.72, 0.85] & 0.83 [0.75, 0.88] \\
\hline
\end{tabular}

\end{minipage}%
}
\end{table}

\textbf{Internal diagnostic audit (ours only).} A blinded post-hoc audit of our predictions was conducted using the predefined rubric described in \textbf{Appendix F \cite{ref38}}:

C.1 GT-Confirmed (same canonical entity),

C.2 GT-Refinement (ontology-consistent and more appropriate for the header),

C.3 Ontology-Consistent Alternative (plausible but not in GT), and C.4 Incorrect.

\textbf{Panel B of Table 7} reports the cumulative diagnostics (B1 strict + C.1, B2 + C.2, B3 + C.3) together with \textbf{95\% CI}. These figures are not revised benchmark scores and are not directly comparable to the official GT-strict results in \textbf{Panel A}. Here, C.2 GT-Refinement contributed \textbf{43/141 (30.5\%)} cases and C.3 Alternatives \textbf{6/141 (4\%)}. 
Cumulative diagnostic support rises from 0.45 under strict GT to ≈ 0.75 with C.2, and to ≈ 0.79 with C.3. This suggests that a substantial subset of GT-strict mismatches is associated with label granularity, aliasing, or ontology-selection effects, rather than with wholly implausible header-centric predictions. This audit, therefore, highlights recurring \textbf{SemTab-2024 GT} limitations under metadata-only evidence and supports the framework's use for KG-oriented benchmark diagnosis and quality analysis.

\subsection{Illustrative Case Studies from Real Datasets}
\label{sec:dq_case_studies}

To demonstrate the operational behaviour of the framework, we highlight representative examples from real-world datasets.

\paragraph{Dataset 352 – Online Retail.}

The \texttt{Description} attribute, classified as \texttt{String}, triggered:

\begin{itemize}
\item DQI \#1 (Missing Data): 1,454 blank or null values.
\item DQI \#17 (Wrong Data Type): one numeric value incorrectly present in a textual field.
\end{itemize}

The \texttt{CustomerID} attribute, classified as \texttt{IDcolumn}, revealed layered violations:

\begin{itemize}
\item DQI \#1 (Missing Data): 135,080 missing values.
\item DQI \#9 (Duplicates): 541,830 repeated identifiers.
\item DQI \#19 (Uniqueness Violation): 537,536 explicit uniqueness violations.
\end{itemize}

This illustrates how identifier semantics activate simultaneous completeness and uniqueness constraints.

\paragraph{Dataset 45 – Heart Disease.}

The attribute \texttt{thalach} (maximum heart rate), classified as numerically bounded (\textit{\textbf{FinalFormat}} \textit{heartrate}), triggered a domain violation of DQI \#15 due to a value exceeding the medically plausible upper bound. 

In contrast, the attribute \texttt{age}, also a numerically bounded type (\textit{\textbf{FinalFormat}} \textit{age}), passed all validation checks with observed values within the expected range (29–77), demonstrating the correct semantic alignment.

\subsection{ \textit{\textbf{HeadersIQ}}: Data Source-Level Data Quality Measurement}
\label{sec:headersiq_results}

Applying the metric defined in Subsection~\ref{sec:HeadersIQ}, we compute data source-level quality scores across the evaluated benchmarks.

For example, Dataset 45 (303 rows, 14 columns; $T = 4242$) contains $V = 7$ cells with at least one detected violation:

\[
\textit{\textbf{HeadersIQ}} = \left(1 - \frac{7}{4242}\right) \times 100 \approx 99.83
\]

Scores close to 100 indicate few detected violations, while lower values reflect increasing levels of detected structural, semantic, or representational issues. Because the metric is normalised by data source size, it enables direct comparison across heterogeneous tabular sources. 

\textit{\textbf{HeadersIQ}} supports quality-aware data preparation by enabling ranking and monitoring of tabular sources prior to knowledge graph integration. In this sense, it provides a lightweight governance signal relevant to data source-level reliability in settings such as \textbf{SemTab 2024 “IsGold?”} and related KG quality assessment work, including \textbf{KG2Tables}.

Representative examples and the full results table are provided in \textbf{Appendix G \cite{ref38}}, together with implementation details in the \textit{Attribute-BasedDataQualityAssessment.ipynb} notebook \cite{ref38}.

\section{Discussion}

\subsection{Reassessing Header-Centric Type Detection}

Header-centric annotation has often been regarded as unreliable, partly influenced by early approaches such as Sherlock \cite{ref13} and Sato \cite{ref45}. Our results challenge this view. With systematic preprocessing, curated and incrementally enriched dictionaries, and abbreviation expansion, headers alone contain enough semantic structure to support precise and scalable type assignments. The 39 \textit{\textbf{FinalFormat}} types provide a compact operational layer between primitive data types and fine-grained ontology classes: richer than atomic types, more compact than the Sherlock/Sato label space, and simple enough to trigger interpretable data quality rules, while \textit{\textbf{SourceKeywords}} preserve finer-grained lexical evidence for auditing and KG-oriented interpretation.

These comparisons should be interpreted in light of task setting. Systems such as Sherlock and Sato were designed primarily for broader semantic type detection scenarios that may exploit richer supervision or table-value context, whereas the present framework targets metadata-only, header-centric interpretation with explicit traceability and DQA activation. The goal is therefore not to outperform value-aware systems on their native setting, but to show that interpretable header-centric typing remains practically effective at scale under metadata-only constraints.

\subsection{General-Purpose Core and Domain-Specific Long-Tail Modelling}

A modest dictionary expansion (approximately 80\% more entries) enabled more than 100-fold growth in practical coverage, increasing valid assignments from approximately 1,000 in the UCI phase to roughly 103,000 across more than 118,000 heterogeneous headers. This suggests that NIL-driven feedback can expand lexical coverage substantially without requiring equally large growth in the dictionaries. At the same time, the 76 Sherlock/Sato types map cleanly into only 15 core \textit{\textbf{FinalFormat}} categories. An additional 24 domain-specific types extend the taxonomy to cover real-world long-tail cases without creating an overly fragmented label space. Consistent results across UCI, Prague, Sato/VizNet, Kaggle, SOTAB, and T2Dv2 indicate that the same set of 39 \textit{\textbf{FinalFormats}} remains broadly usable across heterogeneous benchmarks, while the parallel KG-mapping pathway maintains alignment with DBpedia and Schema.org. In this sense, the 39 \textit{\textbf{FinalFormats}} function as a compact operational vocabulary: large enough to support interpretable DQA activation and real-world long-tail cases, yet small enough to remain auditable, stable, and reusable across diverse data sources.

\subsection{Data Source-Level Quality via \textit{\textbf{HeadersIQ}}}

\textit{\textbf{HeadersIQ}} provided intuitive data quality scores across sources of varying sizes, offering a scalable way to monitor detected issues in large repositories. Its design is relevant to \textbf{SemTab 2024 “IsGold?”} and \textbf{KG2Tables}-style data source reliability questions and illustrates the practical value of linking semantic typing with automated data-quality assessment.

\subsection{Interpreting Ground Truth in SemTab 2024}

Official \textbf{SemTab 2024 Metadata-to-KG} results yielded Hit@1 = 0.45 and Hit@5 = 0.48. A blinded header-centric audit indicates that a substantial subset of GT-strict mismatches is associated with granularity, aliasing, or ontology-selection effects. This diagnostic evidence suggests that rigid GT matching can obscure semantically plausible metadata-only predictions and supports future CTA benchmarks with multi-label or graded evaluation, together with explicit human-in-the-loop validation.

\subsection{Limitations and Future Work}

The main limitations are (i) dependence on meaningful headers and (ii) difficulty with long-tail terms. Several results in this paper are coverage-oriented rather than direct correctness measures; ongoing work includes a \textbf{Statistical Audit} for \textit{\textbf{FinalFormat}} correctness and a KG vocabulary-gap analysis of frequent \textit{\textbf{SourceKeywords}} without exact DBpedia or Schema.org terms. Future work should combine header and value signals, integrate multilingual models, and explore LLM-based disambiguation for sparse, overloaded, or over-specialised metadata

\section{Conclusion}

This paper presents an explainable, header-centric CTA framework for metadata-only semantic typing that combines normalization, curated dictionaries, and \textit{\textbf{SourceKeywords}} to assign 39 \textit{\textbf{FinalFormat}} types. The same abstraction layer supports mappings to DBpedia and Schema.org, grounding cell-level data quality checks that aggregate into the data source-level score \textit{\textbf{HeadersIQ}}.

Experiments on around 120,000 header columns from seven benchmarks, together with a diagnostic audit of the \textbf{SemTab 2024 Metadata-to-KG track}, show that the framework supports large-scale metadata-only semantic table interpretation, while also providing useful benchmark-diagnostic evidence for KG-oriented evaluation. This positions the framework as a practical contribution to metadata-driven semantic annotation, quality-aware tabular preparation, and trustworthy knowledge-graph integration. More broadly, the framework also reflects robustness concerns that remain central in recent SemTab editions, particularly under noisy, ambiguous, and incomplete tabular metadata, while also opening opportunities for multilingual and LLM-assisted extensions in future work.

\section*{Declaration on Generative AI}

During the preparation of this work, the authors used ChatGPT/GPT5 (OpenAI) in order to: Grammar and spelling check, Paraphrase and reword, Improve writing style, and Formatting assistance. Further, the authors used ChatGPT/GPT5 (OpenAI) to assist with the development, debugging, and explanation of code used in the research workflow. After using this tool, the authors reviewed and edited the content as needed and take full responsibility for the publication's content.

\bibliography{references}

\end{document}